%% file: aspdac_LanCe.tex
\documentclass[twocolumn]{article}
\advance\topmargin-1in\advance\oddsidemargin-1in
\evensidemargin\oddsidemargin

\makeatletter
\def\@normalsize{\@setsize\normalsize{12pt}\xpt\@xpt
\abovedisplayskip 10pt plus2pt minus5pt\belowdisplayskip \abovedisplayskip
\abovedisplayshortskip \z@ plus3pt\belowdisplayshortskip 6pt plus3pt
minus3pt\let\@listi\@listI}

\def\section{\@startsection {section}{1}{\z@}{20pt plus 2pt minus 2pt}
{8pt plus 2pt minus 2pt}{\centering\normalsize\sc
\edef\@svsec{\thesection.\ }}}
\def\thesection{\Roman{section}}

\def\subsection{\@startsection {subsection}{2}{\z@}{16pt plus 2pt minus 2pt}
{6pt plus 2pt minus 2pt}{\normalsize\sl
\edef\@svsec{\thesubsection.\ }}}
\def\thesubsection{\Alph{subsection}}

\long\def\@makecaption#1#2{
\vskip10pt\begin{center} #1 #2 \end{center}\par\vskip 1pt}
\def\fnum@figure{\raggedright{\footnotesize Fig. \thefigure }.%
\footnotesize}
\def\fnum@table{\footnotesize TABLE \thetable\\\footnotesize\sc}
\def\thetable{\Roman{table}}

\makeatother
\usepackage{booktabs}
\usepackage{algorithm}
\usepackage{algpseudocode}
\usepackage{amsmath}
\usepackage{amssymb}
\usepackage{array}
\usepackage{balance}
\usepackage[pangram]{blindtext}
\usepackage[noadjust]{cite}
\usepackage{caption}
\usepackage{color}
\usepackage{comment}
\usepackage{dblfloatfix}
\usepackage[inline]{enumitem}
\usepackage{epsfig}
\usepackage{etoolbox}
\usepackage{fancybox}
\usepackage{filecontents}
\usepackage{fixltx2e}
\usepackage{float}
\usepackage{graphicx}
\usepackage[bookmarks=false]{hyperref}
\usepackage{hyperref}
\usepackage{lastpage}
\usepackage{lipsum}
\usepackage{listings}
\usepackage{multirow}
\usepackage{multicol}
\usepackage{pifont}
\usepackage{placeins}
\usepackage{rotating}
\usepackage{setspace}
\usepackage{soul}
\usepackage[hang, tight]{subfigure}
\usepackage{threeparttable}
\usepackage{times}
\usepackage{url}
\usepackage{verbatim}
\usepackage{wrapfig}
\usepackage[usenames,dvipsnames]{xcolor}
\usepackage{color}

\newcolumntype{I}{!{\vrule width 3pt}}
\newlength\savedwidth

\newlength\savewidth

\algtext*{EndWhile}
\algtext*{EndIf}
\algtext*{EndFor}
\algtext*{EndProcedure}

\begin{document}
\date{}

\title{
\vspace{-26mm}
\Large\textbf{LanCe: A Comprehensive and Lightweight CNN Defense Methodology against Physical Adversarial Attacks on Embedded Multimedia Applications}
\vspace{-4mm}}	


\author{Zirui Xu, Fuxun Yu, Xiang Chen\\
George Mason University, Fairfax, Virginia \\
{\small $\{$zxu21, fyu2, xchen26$\}$@gmu.edu }
\vspace{-4mm}}

\maketitle
\thispagestyle{empty}


\newbool{inccomment}
\setbool{inccomment}{true}
\newcommand{\XX}[1]{\ifbool{inccomment}{{\color{magenta} #1}}{}}
\newcommand{\CT}[1]{\ifbool{inccomment}{{\color{magenta}CT\@: #1}}{}}
\newcommand{\NT}[1]{\ifbool{inccomment}{{\color{blue}NT\@: #1}}{}}
\newcommand{\TD}[1]{\ifbool{inccomment}{{\color{orange}#1}}{}}
\newcommand{\FN}[1]{\ifbool{inccomment}{{\color{OliveGreen}#1}}{}}
\newcommand{\GR}[1]{\ifbool{inccomment}{{\color{Tan}#1}}{}}
\newcommand{\LD}{\ifbool{inccomment}{{\color{magenta}\\============================================\\}}}
\newcommand{\RF}{\ifbool{inccomment}{{\color{green}~[R]}}}
\newcommand{\tabincell}[2]{\begin{tabular}{@{}#1@{}}#2\end{tabular}}
\newcommand{\roma}[1]{\uppercase\expandafter{\romannumeral #1\relax}}
\graphicspath{}

\input{0_abstract}

\input{1_introduction}
\input{2_preliminary}
\input{3_theory}

\input{4_image}
\input{5_audio}

\input{6_experiment}

\input{7_conclusion}

\let\oldbibliography\thebibliography
\renewcommand{\thebibliography}[1]{%
  \oldbibliography{#1}%
  \setlength{\itemsep}{-5pt}%
}

\vspace{-7mm}
\scriptsize
\bibliographystyle{IEEEtran}
\bibliography{KDD}

\end{document}

%% file: 0_abstract.tex
\begin{abstract}

Recently, adversarial attacks can be applied to the physical world, causing practical issues to various Convolutional Neural Networks (CNNs) powered applications.
Most existing physical adversarial attack defense works
 only focus on eliminating explicit perturbation patterns from inputs, ignoring interpretation to CNN's intrinsic vulnerability.
	Therefore, they lack expected versatility to different attacks and thereby depend on considerable data processing costs. 
In this paper, we propose \textit{LanCe} -- a comprehensive and lightweight CNN defense methodology against different physical adversarial attacks.
	By interpreting CNN's vulnerability, we find that non-semantic adversarial perturbations can activate CNN with significantly abnormal activations and even overwhelm other semantic input patterns' activations.
		We improve the CNN recognition process by adding a self-verification stage to detect the potential adversarial input with only one CNN inference cost.
	Based on the detection result, we further propose a data recovery methodology to defend the physical adversarial attacks.
	We apply such defense methodology into both image and audio CNN recognition scenarios
	and analyze the computational complexity for each scenario, respectively.
Experiments show that our methodology can achieve an average 91\% successful rate for attack detection and 89\% accuracy recovery. Moreover, it is at most 3$\times$ faster compared with the state-of-the-art defense methods, making it feasible to resource-constrained embedded systems, such as mobile devices.

\end{abstract}

%% file: 1_introduction.tex
\vspace{-5mm}
\section{Introduction}
\label{sec:intr}
\vspace{-2mm}

\begin{figure}[t]
	\centering
	\captionsetup{justification=centering}
	\vspace{1mm}
	\includegraphics[width=3.3in]{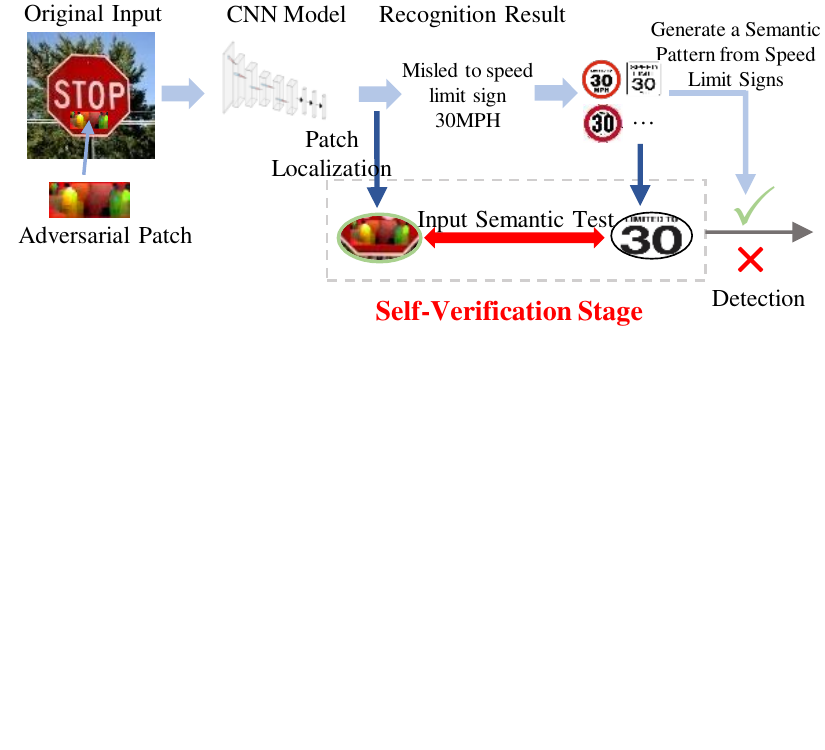}
	\vspace{-45mm}
	\caption{Physical Adversarial Attack for Traffic Sign }
	\vspace{-6mm}
	\label{Threat_Model}
\end{figure}

In the past few years, Convolutional Neural Networks (CNNs) powered applications are facing a critical challenge -- adversarial attacks.
	By injecting particular perturbations into input data, adversarial attacks can mislead CNN recognition results.
	With aggressive methods proposed, adversarial perturbations can be concentrated into a small area and attached to the real objects, which easily threaten the CNN recognition systems in the physical world.
The left side of Fig.~\ref{Threat_Model} shows a physical adversarial example on the traffic sign detection.
	When attaching a well-crafted adversarial patch on the original stop sign, the traffic sign detection system will be misled to a wrong recognition result as a speed limit sign.


Many works have been proposed to defend against physical adversarial attacks~\cite{hayes2018visible,naseer2019local,yang2018characterizing,multiversion}.
	However, most of them neglected CNN's intrinsic vulnerability interpretations.
	Instead, either they merely focused on eliminating explicit perturbation patterns from input~\cite{naseer2019local}, or they simply adopted multiple CNNs to conduct the cross-verification~\cite{multiversion,yang2018characterizing}.
	All these methods have certain drawbacks:
	They failed to find a common defense methodology, lacking versatility for preventing different physical adversarial attacks.
	Moreover, they introduced considerable data processing costs during perturbations elimination, which significantly increased methods' computation costs.

In this paper, we propose \textit{LanCe}, a comprehensive and lightweight defense methodology against different physical adversarial attacks.
	By interpreting CNN's vulnerability, we reveal that the CNN decision-making process lacks necessary qualitative semantics distinguishing ability: the non-semantic input patterns can significantly activate CNN and overwhelm other semantic input patterns.
Leveraging the adversarial attacks' characteristic inconsistencies, we improve the CNN recognition process by adding a self-verification stage. 
	Fig.~\ref{Threat_Model} illustrates the self-verification stage for a traffic sign adversarial attack.
	For each input image, after one CNN inference, the verification stage will locate the significant activation sources (green circle) and calculate the input semantic inconsistency with the expected semantic patterns (right circle) according to the prediction result. 
	Once the inconsistency exceeds a pre-defined threshold, CNN will conduct a data recovery process to recover the input image.
	Our defense methodology has minimum computation components involved, which can be extended to CNN based image and audio recognition scenarios.

Specifically, we have following contributions in this work:
\begin{itemize}
	\vspace{-2.5mm}
	\item By interpreting CNN's vulnerability, we identify characteristic inconsistencies between the physical adversarial attack and the natural input recognition.
	\vspace{-2.5mm}
	\item We propose a self-verification stage to detect the abnormal activation patterns' semantics with only one CNN inference involved. 
	\vspace{-2.5mm}
	\item We further propose a data recovery methodology to recover both attacked image and audio input data. Moreover, we apply such detection and data recovery methodology into image and audio scenarios.
	\vspace{-2.5mm} 
	\item In each scenario, we quantitatively analyze our defense process's computational complexity, and guarantee the lightweight computation cost. 
		\vspace{-2mm}
\end{itemize}

			\vspace{-1mm}
Experiments show that our method can achieve an average 90\% detection successful rate and average 81\% accuracy recovery for image physical adversarial attacks.
Also, our method achieves 92\% detection successful rate and 77.5\% accuracy recovery for audio adversarial attacks.
Moreover, our method is at most 3$\times$ faster than the state-of-the-art defense methods, which is feasible to various resource-constrained embedded systems, such as mobile devices.

%% file: 2_preliminary.tex
\vspace{-5mm}
\section{Background and Related Works}
\label{sec:prel}
\vspace{-1mm}

\subsection{\textbf{Physical Adversarial Attacks}}
\vspace{-2mm}

\begin{figure}[t]
	\centering
	\captionsetup{justification=centering}
	\vspace{-18mm}
	\includegraphics[width=3.3in]{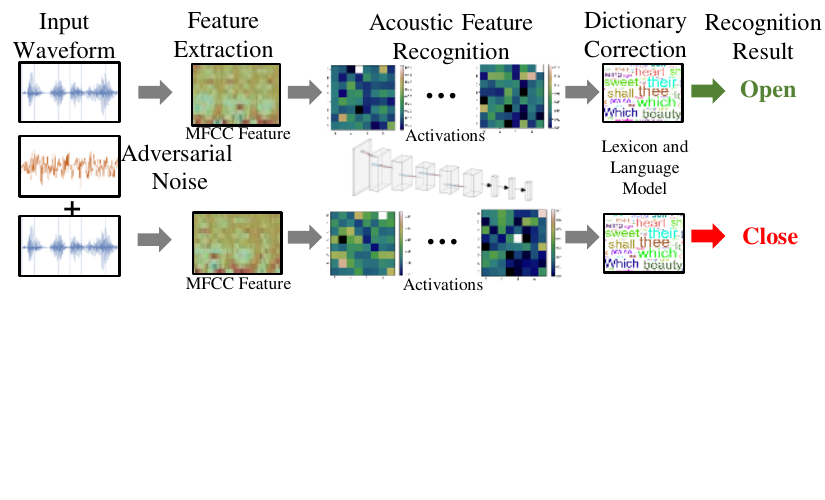}
	\vspace{-22mm}
	\caption{Audio Recognition and Physical Adversarial \\Attack Process}
	\vspace{-6mm}
	\label{Audio_Adv}
\end{figure}

Adversarial attacks started to arouse researchers' general concern with adversarial examples, which were first introduced by~\cite{goodfellow2014explaining}.
Recently, adversarial attack approaches were also brought from the algorithm domain into the physical world, which are referred as the physical adversarial attack.
	~\cite{eykholt2017robust} first leveraged a masking method to concentrate the adversarial perturbations into a small area and implement the attack on real traffic signs with taped graffiti.
	~\cite{brown2017adversarial} extended the scope of physical attacks with adversarial patches.
	With more aggressive patterns than graffiti, these patches can be attached to physical objects arbitrarily and have strong model transferability.

Beyond aforementioned image cases, some physical adversarial attacks also have been proposed to audios. 
	Yakura \textit{et al.}~\cite{yakura2018robust} proposed an audio physical adversarial attack that can still be effective after playback and recording in the physical world.
	~\cite{commandersong} generated audio adversarial commands in a normal song which can be played through the air.

	Compared to noise based adversarial attacks, physical adversarial attacks reduce the attack difficulty and further impair the practicality and reliability of deep learning technologies.

\vspace{-4mm}
\subsection{\textbf{Image physical Adversarial Attack Defense}}
\vspace{-2mm}

There are several works have been proposed to defense such physical adversarial attacks in the image recognition process.
	Naseer \textit{et al.} proposed a local gradients smoothing scheme against physical adversarial attacks~\cite{naseer2019local}. By regularizing gradients in the estimated noisy region before feeding images into CNN inference, their method can eliminate the potential impacts from adversarial attacks.
	Hayes \textit{et al.} proposed a physical image adversarial attack defense method based on image inpainting~\cite{hayes2018visible}. Based on the traditional image processing methods, they detect the localization of adversarial noises in the input image and further leverage the image inpainting technology to remove the adversarial noises.

Although these methods are effective for image physical adversarial attacks defense, they still have certain disadvantages regarding versatility and computation. 
These methods are designed for solving specific adversarial attack which are not integrated for different physical adversarial attack situations. Moreover, they will introduce huge computation costs.

\vspace{-4mm}
\subsection{\textbf{Audio Physical Adversarial Attack Defense}}
\vspace{-2mm}


Compared with images, the audio data requires more processing efforts for recognition.
	Fig.~\ref{Audio_Adv} shows a typical audio recognition process and the corresponding physical adversarial attack.
	The audio waveform is first extracted as Mel-frequency Cepstral Coefficient (MFCC) features.
	Then we leverage a CNN to achieve acoustic feature recognition, which can obtain the candidate phonemes.
	Finally, a lexicon and language model is applied to obtain the recognition result "open".
When the adversarial noise is injected to the original input waveform, 
the final recognition result is misled to "close".

Several works have been proposed to detect and defend such adversarial attacks~\cite{multiversion, yang2018characterizing,rajaratnam2018noise}.
	Zeng \textit{et al.} leveraged multiple Automatic Speech Recognition (ASR) systems to detect audio physical adversarial attack based on a cross-verification methodology~\cite{multiversion}. However, their method lacks certain versatility which cannot detect the adversarial attacks with model transferability.
	Yang \textit{et al.} proposed an audio adversarial attack detection and defense method by exploring the temporal dependency in audio adversarial attacks~\cite{yang2018characterizing}.
	However, their method requires multiple CNN recognition inferences which is time-consuming.


%% file: 3_theory.tex
\vspace{-5mm}
\section{Interpretation Oriented Physical Adversarial Attacks Analysis and Defense}
\label{sec:algo}
\vspace{-1.5mm}



In this section, we first interpret the CNN vulnerability by analyzing input patterns' semantics with the activation maximization visualization~\cite{erhan2009visualizing}. 
Based on semantics analysis, we identify the adversarial attack patches as non-semantic input patterns with abnormal distinguished activations. 
Specifically, to evaluate the semantics, we propose metrics that can measure inconsistencies between the local input patterns that cause the distinguished activations and the synthesized patterns with expected semantics.
Based on the inconsistency analysis, we further propose a lightweight defense methodology consists of the self-verification and the data recovery. 

\begin{figure}[t]
	\centering
	\captionsetup{justification=centering}
	\vspace{-17mm}
	\includegraphics[width=3.3in]{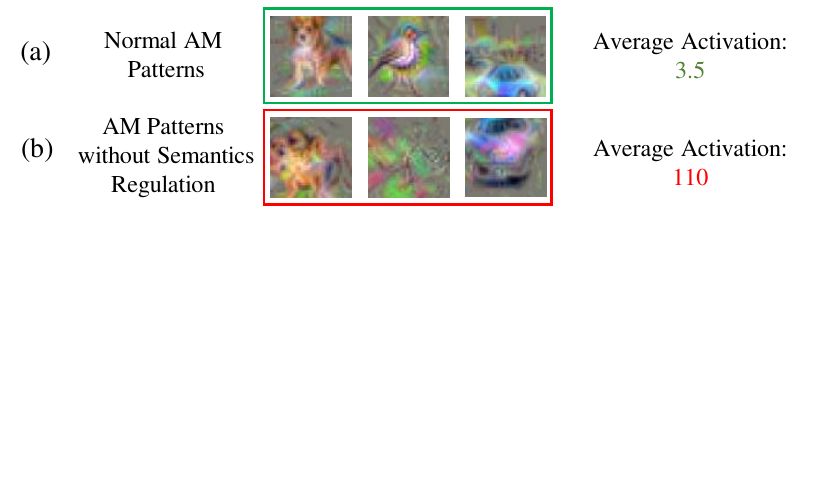}
	\vspace{-30mm}
	\caption{Visualized Neuron's Input Pattern by Activation Maximization Visualization}
	\vspace{-6mm}
	\label{Semantic_Mismatch}
\end{figure}

\vspace{-4mm}
\subsection{\textbf{CNN Vulnerability Interpretation}}
\vspace{-2mm}

\textit{\textbf{Interpretation and Assumption:}}
In a typical image or audio recognition process, CNN extracts features from the original input data and gradually derive a prediction result.
However, when injecting physical adversarial perturbations into the original data, CNN will be misled to a wrong prediction result. 
To better interpret the vulnerability, we major focus on a typical image physical adversarial attack -- adversarial patch attack as an example.
In Fig.~\ref{Threat_Model}, by comparing with the original input, we find that an adversarial patch usually has no constraints in color/shape, \textit{etc}. 
Such patches usually sacrifice the semantic structures so as to cause significant abnormal activations and overwhelm the other input patterns' activations. 
\textit{Therefore, we make an \textbf{assumption} that CNN lacks qualitative semantics distinguishing ability which can be activated by the non-semantic adversarial patch during CNN inference.}

\textit{\textbf{Assumption Verification:}}
	According to our assumption, the non-semantic input patterns will lead to abnormal activations while the semantic input patterns generate normal activations.
We can evaluate this difference by investigating the semantic of each neuron in CNN. 
Therefore, we adopt a visualized CNN semantic analysis method -- Activation Maximization Visualization (AM)~\cite{erhan2009visualizing}. 
	AM can generate a pattern to visualize each neuron's most activated semantic input. 
The generation process of pattern $V(N_i^l)$ can be considered as synthesizing an input image to a CNN model that delicately maximizes the activation of the $ith$ neuron $N_i^l$ in the layer of $l$. Specifically, this process can be formulated as:
\vspace{-2mm}
\small
\begin{equation}
	\medmuskip=-1mu
	V(N_i^l)=\mathop{\arg\max}_{X} A_i^l(X), \qquad X \leftarrow X + \eta \cdot \frac{\partial A_i^l(X)}{\partial X}
	\label{eq:am}
	\vspace{-2mm}
\end{equation}
\normalsize
where, $A_i^l(X)$ is the activation of $N_i^l$ from an input image X, $\eta$ is the gradient ascent step size.

Fig.~\ref{Semantic_Mismatch} shows the visualized semantic input patterns by using AM. 
As the original AM method is designed for semantics interpretation, many feature regulations and hand-engineered natural image references are involved in generating interpretable visualization patterns. 
Therefore we can get three AM patterns with an average activation magnitude value of 3.5 in Fig.~\ref{Semantic_Mismatch} (a). 
The objects in the three patterns indicate they have clear semantics. 
However, when we remove these semantics regulations in the AM process, we obtain three different visualized patterns as shown in Fig.~\ref{Semantic_Mismatch} (b). 
We can find that these three patterns are non-semantic, but they have significant abnormal activations with an average magnitude value of 110.  
This phenomenon can prove our assumption that CNN neurons lack semantics distinguishing ability and can be significantly activated by \textit{\textbf{non-semantic}} inputs patterns. 

\begin{figure}[t]
	\centering
	\captionsetup{justification=centering}
	\vspace{-21mm}
	\includegraphics[width=3.1in]{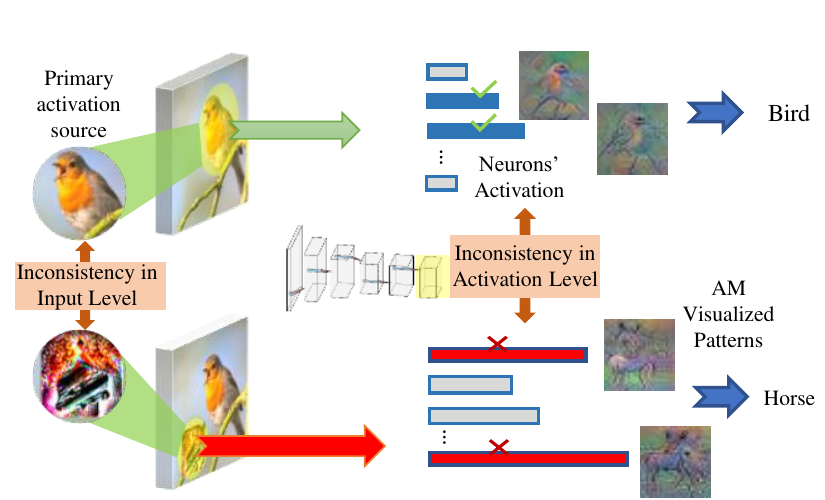}
	\vspace{-3mm}
	\caption{Image Adversarial Patch Attack}
	\vspace{-6mm}
	\label{Adversarial_Patch}
\end{figure}

\vspace{-4mm}
\subsection{\textbf{Inconsistency Metrics for Input Semantic \\and Prediction Activation}}
\vspace{-2mm}

\textit{\textbf{Inconsistency Identification:}}
To identify the non-semantic input patterns for the attack detection, we examine its impacts during CNN inference by comparing the natural image recognition with the physical adversarial attacks. 

Fig.~\ref{Adversarial_Patch} shows a typical adversarial patch based physical attack.
The patterns in the left circles are the primary activation sources from the input images, and the bars on the right are the neurons' activations in the last convolutional layer.
From input patterns, we identify a significant difference between the adversarial patch and primary activation source on the original image, which is referred as \textit{\textbf{Input Semantic Inconsistency}}.
From the aspect of prediction activation magnitudes, we observe another difference between the adversarial input and the original input, namely \textit{\textbf{Prediction Activation Inconsistency}}.

\textit{\textbf{Inconsistency Metrics Formulation:}} We further define two metrics to indicate above two inconsistencies' degrees. 

\textit{\textbf{1) Input Semantic Inconsistency Metric:}}
This metric measures the input semantic inconsistency between the non-semantic adversarial patches and the semantic local input patterns from the natural image. 
It can be defined as follows: 
\vspace{-1.5mm}
\small
\begin{equation}
	\medmuskip=-3mu
	\thinmuskip=-3mu
	\thickmuskip=-3mu
	D(P_{pra},P_{ori})=1-S(P_{pra},P_{ori}), P_{pra} \xleftarrow{\Re} \Phi:{A_i^l(p)}, P_{ori} \xleftarrow{\Re} \Phi:{A_i^l(o)},
	\label{eq:metric1}
	\vspace{-1.5mm}
\end{equation}
\normalsize
where $P_{pra}$ and $P_{ori}$ represent the input patterns from the adversarial input and the original input. 
$\Phi:{A_i^l(p)}$ and $\Phi: {A_i^l(o)}$ represent the set of neurons' activations produced by the adversarial patch and the original input, respectively. $\Re$ maps neurons' activations to the primary local input patterns.  
$S$ represents a similarity metric. 

\textit{\textbf{2) Prediction Activation Inconsistency Metric:}}
The second inconsistency is on the activation level, which reveals the activations' magnitude distribution inconsistency in the last convolutional layer between the adversarial input and the original input. 
We also use a similar metric to measure it as follows: 
\vspace{-1.5mm}
\small
\begin{equation}
	\medmuskip=-1mu
	\thinmuskip=-1mu
	\thickmuskip=-1mu
	D(f_{pra},f_{ori})=1-S(f_{pra},f_{ori}), f_{pra} \sim \Phi:{A_i^l(p)}, f_{ori} \sim \Phi:A_i^l(o),
	\label{eq:metric1}
	\vspace{-0.5mm}
\end{equation}
\normalsize
where $f_{pra}$ and $I_{ori}$ represent the magnitude distribution of activations in the last convolutional layer generated by the adversarial input and the original input data.

For the above two inconsistency metrics, we can easily obtain $P_{pra}$ and $f_{pra}$ since they come from the input data. 
However, $P_{ori}$ and $f_{ori}$ are not easily to get because of the variety of the natural input data. 
Therefore, we need to synthesize the standard input data which can provide the semantic input patterns and activation magnitude distribution.   
The synthesized input data for each prediction class can be obtained from a standard dataset. 
By feeding CNN with a certain number of input from the standard dataset, we can record the average activation magnitude distribution in last convolutional layer. 
Moreover, we can locate the primary semantic input patterns for each prediction class. 

\begin{figure}[t]
	\centering
	\captionsetup{justification=centering}
	\vspace{-18mm}
	\includegraphics[width=3.1in]{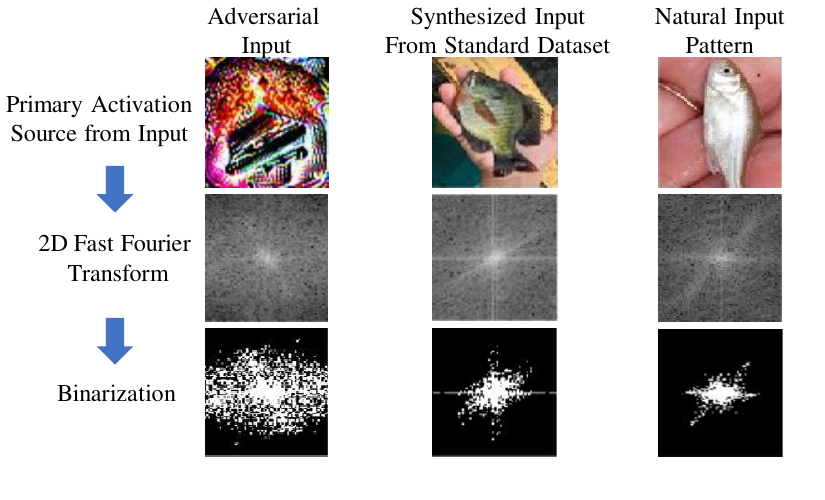}
	\vspace{-6.5mm}
	\caption{The Results after 2D Fast Fourier Transform}
	\vspace{-5.5mm}
	\label{2D_F}
\end{figure}

\vspace{-4mm}
\subsection{\textbf{Physical Adversarial Attack Defense based on \\CNN Self-Verification and Data Recovery}}
\vspace{-2mm}

The proposed two inconsistencies demonstrate the difference between physical adversarial attacks and natural image recognition \textit{w.r.t} input patterns and prediction activations. 
	By utilizing the inconsistency metrics, we propose a defense methodology which consists of a self-verification and a data recovery in the CNN decision-making process. 
Specifically, the entire methodology flow is described as following: 

\textbf{\textit{Self-Verification:}}
(1) We first feed the input into the CNN inference and obtain the prediction class.
(2) Next, CNN can locate the primary activation sources from the practical input and obtain the activations in the last convolutional layer.
(3) Then CNN leverages the proposed metrics to measure the two inconsistencies between the practical input and the synthesized data with the prediction class.
(4) Once any inconsistency exceeds the given threshold, CNN will consider the input as an adversarial input.

\textbf{\textit{Data Recovery:}}
(5) After a physical adversarial attack has been detected by the self-verification stage, the data recovery methodology is further applied to recover the input data which has been attacked. 
Specifically, we leverage image inpainting and activation denoising to recover the input image and audio.

We will derive two methods from such methodology for image and audio scenarios in Section 4 and Section 5. 

\textbf{\textit{Computational Complexity:}}
As aforementioned, the computation cost is critical to the adversarial defense approaches. 
Therefore, we leverage computational complexity to evaluate the methodology's total computation cost. 
	A low computational complexity indicates a small computation workload, proving the proposed methodology is lightweight. 
In our defense methodology, the computational complexity is mainly contributed by the inner steps such as the CNN inference, inconsistency metrics calculation and data recovery.
	In following two scenarios, we will specifically analyze the computation complexity for each of above steps.

%% file: 4_image.tex
\vspace{-5mm}
\section{Defense Against Image Physical \\Adversarial Attack}
\label{sec:imag}
\vspace{-1mm}

In this section, we will specifically describe our defense methodology against image physical adversarial attacks. 


\vspace{-5mm} 
\subsection{\textbf{Defense Process in the Image Scenario}}
\vspace{-2mm} 
\textit{\textbf{Primary Activation Pattern Localization:}}
For the image physical adversarial attacks defense, we mainly depend on the \textit{\textbf{input semantic inconsistency}} in input pattern level. 
Therefore, we need to locate the primary activation source from the input image by adopting a CNN activation visualization method -- Class Activation Mapping (CAM)~\cite{zhou2016learning}. 
Let $A_k(x,y)$ denotes the value of the $k^{th}$ activation in the last convolutional layer at spatial location $(x,y)$. 
	We can compute a sum of all activations at the spatial location $(x,y)$ in the last convolutional layer as:
	\vspace{-2mm} 
\small
\begin{equation}
	\medmuskip=-1mu
	A_{T}(x,y)=\sum^{1}_{K}A_k(x,y),
	\label{eq:cam}
	\vspace{-1.5mm}
\end{equation}
\normalsize
where $K$ is the total number of activations in the last convolutional layer. 
The larger value of $A_{T}(x,y)$ indicates the activation source in the input image at the corresponding spatial location $(x,y)$ is more important for classification result. 





\textit{\textbf{Inconsistency Derivation:}}
According to our preliminary analysis, the input adversarial patch contains much more high-frequency information than the natural semantic input patterns. 
Therefore, we convert the patterns with a series of transformations which are shown in Fig.~\ref{2D_F}. 
	After the 2D Fast Fourier Transform (2D-FFT) transformation and binary conversion, we can observe the significant difference between adversarial input and semantic synthesized input. 
Therefore, we replace $S(I_{pra},I_{ori})$ with Jaccard Similarity Coefficient (JSC)~\cite{niwattanakul2013using} and propose our image inconsistency metric as: 
\vspace{-1mm}
\small
\begin{equation}
	\medmuskip=-3mu
	\thinmuskip=-3mu
	\thickmuskip=-2.5mu
	\hspace{-2mm}
	D(P_{pra},P_{exp})=1-JSC(P_{pra},P_{exp})=\frac{|P_{pra}\bigcup P_{exp}|-|P_{pra}\bigcap P_{exp}|}{|P_{pra}\bigcup P_{exp}|},
	\label{eq:metric1}
	\vspace{-1mm}
\end{equation}
\normalsize
where $I_{exp}$ is the synthesized semantic pattern with predicted class.
$P_{pra}\bigcap P_{exp}$ means the numbers of pixels where the pixel value of $P_{pra}$ and $P_{exp}$ both equal to 1.

\begin{figure}[t]
	\centering
	\captionsetup{justification=centering}
	\vspace{-22mm}
	\includegraphics[width=3.1in]{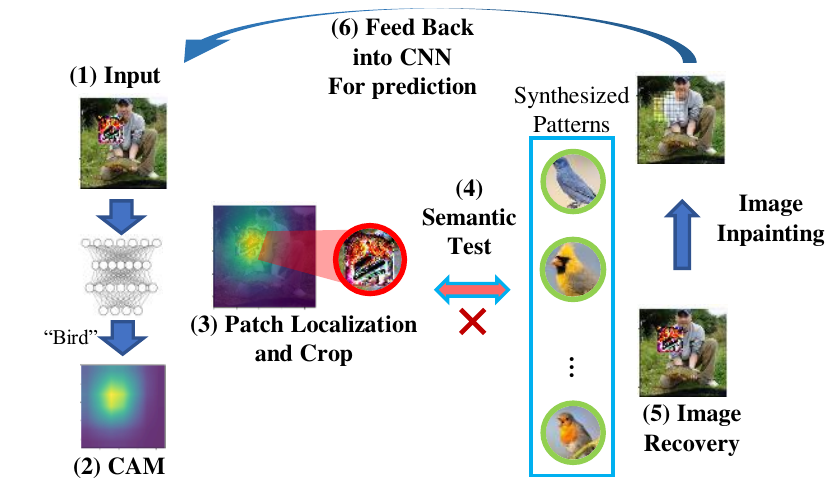}
	\vspace{-4mm}
	\caption{Adversarial Patch Attack Defense}
	\vspace{-6.5mm}
	\label{Image_Patch}
\end{figure}

With the above inconsistency metric, we propose our specific defense methodology which contains self-verification and image recovery.
The entire process is described in Fig.~\ref{Image_Patch}. 

\textbf{\textit{Self-Verification for Detection:}}
For each input image, we apply CAM to locate the source location of the biggest model activations. Then we crop the image to obtain patterns with maximum activations.
During semantic test, we calculate the inconsistency between $I_{pra}$ and $I_{exp}$. 
If it is higher than a predefined threshold, we consider an adversarial input detected.

\textbf{\textit{Data Recovery for Image:}}
After the patch is detected, we conduct the image data recovery by directly removing patch from the original input data.
	In our case, to ensure the lightweight computation workload, we leverage Neighbor Interpolation, a simple but effective image inpainting technology to repair the image and eliminate the attack effects. 
Concretely, each pixel in the adversarial patch will be replace by the average value of its eight surrounding pixels.  
After the interpolation, we feed back the recovery image into CNN to do the prediction again. 
With above steps, we can defend an image physical adversarial attack during CNN inference.

\vspace{-4mm}
\subsection{\textbf{Computational Complexity Analysis}}
\vspace{-2mm}

The total computation complexity of the defense process in the image scenario is contributed by following four steps: the CNN inference, the maximum activation pattern locating, the inconsistency metric calculation and the image interpolation.
	We model each step's computational complexity as following: 

\textit{\textbf{CNN Inference:}}
When the input image is first fed into CNN for class prediction, the inference computational complexity $C_C$ is formulated as:
\vspace{-2mm}
\begin{equation}
\small
	C_C \sim \mathcal{O} (\sum^{L}_{i=1}\sum^{n_i}_{j=1}{r^j_i}^2 n_{i-1} {h^j_i}{w^j_i}),
	\label{eq:metric1}
	\vspace{-2mm}
\end{equation}
\normalsize
where ${r^j_i}^2$ represents $j^{th}$ filter's kernel size in $i^{th}$ layer, $h^j_
i w^j_i$ denotes the corresponding size of output feature map, $L$ is the total layer number and $n_i$ is the filter numbers in $i^{th}$ layer.

\textit{\textbf{Primary Activation Pattern Localization:}} 
Since computation complexities of other operations such as cropping are negligible, we consider CAM contributes the primary computational complexity in this step. 
In CAM, each spatial location $(x,y)$ in the last convolutional layer is the weighted sum of $K$ activations. Therefore, the total computational complexity is: $C_M \sim \mathcal{O}(Kh^{n_L}_L w^{n_L}_L)$, where $h^{n_L}_L w^{n_L}_L$ is the size of the feature map in last convolutional layer.

\textit{\textbf{Inconsistency Metric Derivation:}}
This step consists of 2D-FFT calculation and JSC calculation.
	According to the analysis in~\cite{jsc,fft}, the computational complexities of above two processes can be approximate to $C_F \sim \mathcal{O}(NlogN)$ and $C_J \sim \mathcal{O}(n_alogn_a)$, where $N$ and $n_a$ represent $N$ pixel number in input image and maximum activation pattern, respectively.

\textit{\textbf{Image Interpolation:}} 
For each pixel, the total operation number during interpolation is nine (eight adding operation and one dividing operation). Therefore, the total interpolation computation complexity for the entire adversarial patch is $C_L \sim \mathcal{O}(9n_a)$.

Comparing with the last three steps, the computational complexity of CNN inference dominates the entire computational complexity of our defense methodology in the image scenario. 
	Since our methodology only involves one CNN inference, it usually has less computation cost than other methods.

\textit{\textbf{Case Study:}}
To examine the lightweight of our method, we use VGG-16~\cite{Simo:2014:arXiv} with 224$\times$224 input image as an example.
	According to the built models, the total computation complexity of our defense method is approximate to $ \mathcal{O} (15300M)$ FLOPs (Floating Point Operations) while ~\cite{naseer2019local} is approximate to $ \mathcal{O} (18300M)$. 
	Our method's superiority in terms of computational complexity will be further verified by evaluating the process time cost in Section 6. 

%% file: 5_audio.tex
\vspace{-5mm}
\section{Defense Against Audio Physical\\ Adversarial Attack}
\label{sec:audi}
\vspace{-1mm}

In this section, we will introduce the detailed defense design flow for the audio physical adversarial attacks.

\vspace{-5mm}
\subsection{\textbf{Defense Process in the Audio Scenario}}
\vspace{-2mm}

\textit{\textbf{Inconsistency Derivation:}}
	Different from images, the audio data requires more processing efforts. 
	As Fig.~\ref{Audio_Adv} shows, during the audio recognition, the input waveform needs to pass Mel-frequency Cepstral Coefficient (MFCC) conversion to be transferred from the time domain into the time-frequency domain. 
	In that case, the original input audio data will loss semantics after the MFCC conversion. 
	Therefore, we leverage the \textbf{prediction activation inconsistency} to detect the audio physical adversarial attacks.  

More specifically, we measure the activation magnitude distribution inconsistency between the practical input and the synthesized data with the same prediction class. 
We adopt a popular similarity evaluation method - Pearson Correlation Coefficient (PCC)~\cite{benesty2009pearson} and the inconsistency metric is defined as: 
\small
\begin{equation}
	\medmuskip=-3mu
	\thinmuskip=-3mu
	\thickmuskip=-3mu
	D(f_{pra},f_{exp})=1-PCC(f_{pra},f_{exp})=1-\frac{E[(f_{pra}-\mu_{pra})(f_{exp}-\mu_{exp})]}{\sigma_{pra}\sigma_{exp}},
	\label{eq:metric2}
	\vspace{-1.5mm}
\end{equation}
\normalsize
where $I_{pra}$ and $I_{exp}$ represent the activations in the last convolutional layer for both practical input and synthesized input.  
$\mu_a$ and $\mu_o$ denote mean values of $f_{pre}$ and $f_{exp}$, $\sigma_{pra}$ and $\sigma_{exp}$ are standard derivations, and $E$ means the overall expectation. 

\textbf{\textit{Self-Verification for Detection:}}
With established inconsistency metric, we further apply self-verification stage to CNN for the audio physical adversarial attack. 
The detection flow is described as following: 
	We first obtain activations in the last convolutional layer for every possible input word by testing CNN with a standard dataset. 
	Then we calculate the inconsistency value $D(I_{pra},I_{exp})$.
	If the model is attacked by the audio adversarial attack, $D(I_{pra},I_{exp})$ will exceed a pre-defined threshold. 
	According to our preliminary experiments tested with various attacks, $D(I_{pra},I_{exp})$ of an adversarial input is usually larger than 0.18 while a natural input's $D(I_{pra},I_{exp})$ is usually smaller than 0.1. Therefore, there exists a large range for the threshold to distinguish the natural and the adversarial input audios, which can benefit our accurate detection. 

\begin{figure}[t!]
	\centering
	\captionsetup{justification=centering}
	\vspace{-22mm}
	\includegraphics[width=3.1in]{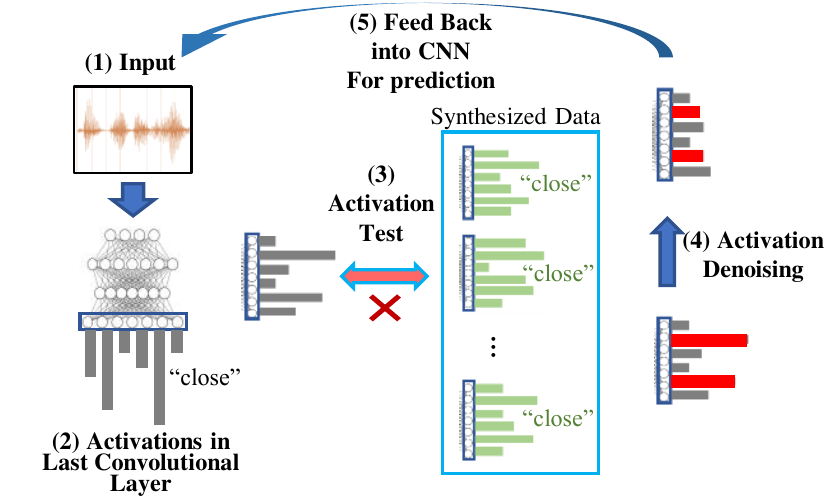}
	\vspace{-4mm}
	\caption{Audio Adversarial Attack Defense }
	\vspace{-7mm}
	\label{Audio_Defense}
\end{figure}

\textbf{\textit{Data Recovery for Audio:}}
After identifying the adversarial input audio, simply denying it can cause undesired consequences. 
	Therefore, attacked audio recovery is considered as one of the most acceptable solutions. 
We propose a new solution - ``activation denoising" as our defense method, which targets ablating adversarial effects from the activation level. 
	The activation denoising takes advantages of the aforementioned last layer activation patterns, which have stable correlations with determined predication labels. 

Our adversarial audio recovery method is shown in Fig.~\ref{Audio_Defense}: 
Based on detection results, we can identify the wrong prediction label, and obtain the standard activation patterns of the wrong class in the last layer. (For the best performance, we locate the top-\textit{k} activation index.)
Then we can find the activations with the same index. These activations are most potentially caused by the adversarial noises and supersede the original activations. 
	Therefore, we suppress these activations to resurrect original ones.

\vspace{-4mm}
\subsection{\textbf{Computational Complexity Analysis}}
\vspace{-2mm}

The computational complexity in the audio scenario is mainly determined by the CNN inference and the inconsistency metric calculation, since other steps directly manipulate limited activation values with negligible computation workload involved.
	Therefore, we model the computational complexity as following:

\textit{\textbf{CNN Inference:}} Since the audio has same inference process in CNN, we can use the same model in image scenario to measure the computation complexity in the audio scenario. 

\textit{\textbf{Inconsistency Metric Derivation:}}
The computation complexity of this step is contributed by the PCC calculation, which can be formulated as $C_P \sim \mathcal{O} ({n_L}^2)$, where $n_L$ is the activation number in the last layer. 

\textit{\textbf{Case Study:}}
We also leverage a case study to specifically demonstrate that our proposed methodology is lightweight comparing with others in the audio scenario.
	The CNN model is Command Classification model~\cite{morgan2001speech} with 1\textit{s} audio input (16000 sample rate). 
Therefore, the total computation complexity of our methodology is approximate to $\mathcal{O} (500M)$ FLOPs (Float Point Operations). However, the computation complexities of other two state-of-the-art audio defense methods are around $ \mathcal{O} (1100M)$ and $ \mathcal{O} (1600M)$. Therefore, our proposed methodology is more friendly to resource-constrained mobile devices.

%% file: 6_experiment.tex
\vspace{-5mm}
\section{Experiment and Evaluation}
\label{sec:Expe}
\vspace{-2mm}

In this section, we evaluate \textit{LanCe} in terms of effectiveness and efficiency for image and audio physical adversarial attacks. 

\vspace{-4mm}
\subsection{\textbf{Defense Evaluation for Image Scenario}}
\vspace{-2mm}

\textit{\textbf{Experiment Setup:}}
Our detection method is mainly evaluated for adversarial patch attacks.
	The adversarial patches are generated by using Inception-V3~\cite{Szeg:2015:CVPR} as the base model. The generated patch with high transferability are utilized to attack other two models: VGG-16~\cite{Simo:2014:arXiv} and ResNet-18~\cite{He:2016:CVPR}.
	Then we apply our defense method on all three models and test their detection and recovery success rates. Meanwhile, we also record the time cost of defense methods to demonstrate the efficiency of \textit{LanCe}. The baseline methods is \textit{Patch Masking}, which is one state-of-the-art defense method~\cite{hayes2018visible}. And the threshold for inconsistency is set as 0.46.

\textit{\textbf{Defense Effectiveness:}}
Table~\ref{tab:2} shows the overall detection and image recovery performance. 
	On all three models, \textit{LanCe} consistently shows higher detection success rate than~\cite{hayes2018visible}.
	The further proposed image recovery could help to correct predictions, resulting in 80.3\%$\sim$82\% accuracy recovery improvement on different models while \textit{Patch Masking} only achieves 78.2\%$\sim$79.5\% accuracy recovery improvement.

\begin{table}[t]
	\vspace{-19mm}
	\centering
	\caption{Image Adversarial Patch Attack Defense Evaluation}
	\vspace{-4mm}
\setlength{\tabcolsep}{0.2mm}{
\footnotesize
\begin{tabular}{|c|c|c|c|c|c|c|c|}
\hline
\multirow{2}{*}{Stage}    & \multirow{2}{*}{}                                             & \multicolumn{2}{c|}{Inception-V3} & \multicolumn{2}{c|}{VGG-16} & \multicolumn{2}{c|}{ResNe-18t} \\ \cline{3-8} 
                          &                                                               & \textit{PM*}       & \textit{LanCe}              & \textit{PM*}   & \textit{LanCe}            & \textit{PM*}      & \textit{LanCe}            \\ \hline
Detection                 & \begin{tabular}[c]{@{}c@{}}Detection\\ Succ.Rate\end{tabular} & 88\%          & \textbf{91\%}       & 89\%      & \textbf{90\%}     & 85\%         & \textbf{89\%}     \\ \hline
\multirow{2}{*}{Recovery} & \begin{tabular}[c]{@{}c@{}}Original\\ Acc.\end{tabular}       & 9.8\%         & 9.8\%               & 9.5\%     & 9.8\%             & 10.8\%       & 9.8\%             \\ \cline{2-8} 
                          & \begin{tabular}[c]{@{}c@{}}Recovery\\ Acc.\end{tabular}       & 88\%          & \textbf{90\%}       & 89.3\%    & \textbf{91.5\%}   & 90\%         & \textbf{91\%}     \\ \hline
                          & Time                                                          & 233ms        & 192ms              & 315ms    & 243ms            & 461ms       & 318ms            \\ \hline
\end{tabular}}
	\begin{tablenotes}
		\scriptsize
		\item[1] *:Patch Masking (PM)~\cite{hayes2018visible}
	\end{tablenotes}
	\label{tab:2}
	\vspace{-6mm}
\end{table}
\normalsize

\textit{\textbf{Time Cost:}}
We leverage the process time cost to represent the method's computational complexity. We can find that the process time cost of our defense method for one physical adversarial attack is from 67\textit{ms}$\sim$71\textit{ms} while the \textit{Patch Masking} is from 132\textit{ms}$\sim$153\textit{ms}.

By the above comparison, we show that our defense method has better defense performance than \textit{Patch Masking} with respect to both effectiveness and efficiency.


\begin{table}[t]
	\vspace{2mm}
	\centering
	\caption{Audio Adversarial Attack Data Recovery Evaluation}
	\vspace{-4mm}
\setlength{\tabcolsep}{0.5mm}{
\footnotesize
\begin{tabular}{|c|c|c|c|c|c|}
\hline
Method                                                         & FGSM         & BIM          & CW          & Genetic      & Time Cost      \\ \hline
No Recovery                                                     & 10\%          & 5\%           & 4\%           & 13\%          & NA             \\ \hline
\begin{tabular}[c]{@{}c@{}}Dependency\\ Detection~\cite{yang2018characterizing} \end{tabular} & 85\%          & 83\%          & 80\%          & 80\%          & 1813ms         \\ \hline
Noise Flooding~\cite{rajaratnam2018noise}                                                 & 62\%          & 65\%          & 62\%          & 59\%          & 1246ms         \\ \hline
\textit{LanCe}                                                           & \textbf{87\%} & \textbf{88\%} & \textbf{85\%} & \textbf{83\%} & \textbf{521ms} \\ \hline
\end{tabular}}
	\label{tab:3}
	\vspace{-6mm}
\end{table}
\normalsize

\vspace{-5mm}
\subsection{\textbf{Defense Evaluation for Audio Scenario}}
\vspace{-2mm}

\textit{\textbf{Experiment Setup:}}
For audio scenario, we use Command Classification Model~\cite{morgan2001speech} on Google Voice Command dataset~\cite{morgan2001speech}.
	The inconsistency threshold for adversarial detection is obtained by the grid search and set as 0.11 in this experiment.
For comparison, we re-implement another two state-of-the-art defense methods: \textit{Dependency Detection}~\cite{yang2018characterizing} and \textit{Multiversion}~\cite{multiversion}. Four methods~\cite{goodfellow2014explaining,kurakin2016adversarial,carlini2017towards,alzantot2018did} are used as attacking methods to prove the generality of our defense method.
Fig.~\ref{Audio_Detection} shows the overall performance comparison. 

\textit{\textbf{Defense Effectiveness:}}
\textit{LanCe} can always achieve more than 92\% detection success rate for all audio physical adversarial attacks. By contrast, \textit{Dependency Detection} achieves 89\% detection success rate in average while \textit{Multiversion Detection} only have average 74\%.
Therefore, \textit{LanCe} demonstrates the best detection accuracy.
Then we evaluate \textit{LanCe}'s recovery performance. The $k$ value in the top-\textit{k} index is set as 6.
Since \textit{Multiversion}~\cite{multiversion} cannot be used to recovery, we re-implement another method, \textit{Noise Flooding}~\cite{rajaratnam2018noise} as comparison. And we use the original vulnerable model without data recovery as the baseline. 
Table~\ref{tab:2} shows the overall audio recovery performance evaluation. After applying our recovery method, the prediction accuracy significantly increase from average 8\% to average 85.8\%, which is 77.8\% accuracy recovery. Both \textit{Dependency Detection} and \textit{Noise Flooding} have lower accuracy recovery rate, which are 74\% and 54\%, respectively.

\textit{\textbf{Time Cost:}}
For defense efficiency, the computational complexity of \textit{LanCe} is much lower than other methods according to our previous analysis.
	As the result, the time cost of our method is 521\textit{ms} while other two methods usually cost more than 1540\textit{ms} for a single physical adversarial attack. Thus, our defense method is 2$\sim$3$\times$ faster than other two methods.

%% file: 7_conclusion.tex
	\vspace{-5mm}
\section{Conclusion}
\label{sec:conc}
	\vspace{-1mm}

In this paper, we propose a CNN defense methodology for physical adversarial attacks for both image and audio recognition applications.
Leveraging the comprehensive CNN vulnerability analysis and two novel CNN inconsistency metrics, our method can effectively and efficiently detect and eliminate the image and audio physical adversarial attacks.
Experiments show that our methodology can achieve an average 91\% successful rate for attack detection and 89\% accuracy recovery.
Moreover, the proposed defense methods are at most 3$\times$ faster compared to the state-of-the-art defense methods, making them feasible to resource-constrained embedded systems, such as mobile devices.



\begin{figure}[t]
	\centering
	\captionsetup{justification=centering}
	\vspace{-21mm}
	\includegraphics[width=3.3in]{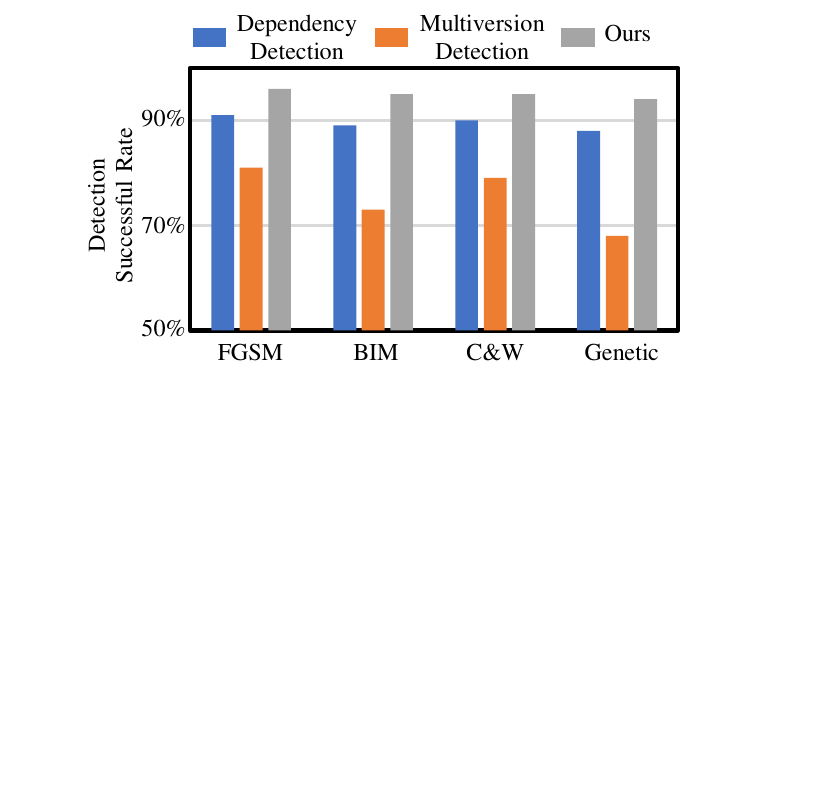}
	\vspace{-48mm}
	\caption{Audio Adversarial Attack Detection Performance}
	\vspace{-5mm}
	\label{Audio_Detection}
\end{figure}